\pdfoutput=1

\documentclass[11pt]{article}

\usepackage[]{acl}

\usepackage{times}
\usepackage{latexsym}

\usepackage[T1]{fontenc}

\usepackage[utf8]{inputenc}

\usepackage{microtype}
\usepackage{multirow}

\usepackage{stfloats}
\usepackage{amsmath}
\usepackage{amssymb}
\usepackage{booktabs}
\usepackage{tabularx}
\usepackage{booktabs}
\usepackage{bm}

\usepackage{adjustbox}

\newcommand\clearrow{\global\let\rowmac\relax}
\title{
How to Distill your BERT: An Empirical Study on the \\Impact of Weight Initialisation and Distillation Objectives
}

\author{
        $\text{Xinpeng Wang}^{*}$\:  $\text{Leonie Weissweiler}^{* \diamond}$ \: $\text{Hinrich Schütze}^{* \diamond}$ \: $\text{Barbara Plank}^{* \diamond}$ \\
        ${}^{*}\text{Center for Information and Language Processing (CIS), LMU Munich, Germany}$  \\
        ${}^{\diamond}\text{Munich Center for Machine Learning (MCML), Munich, Germany}$\\
        {\tt \{xinpeng, weissweiler, bplank\}@cis.lmu.de}}

\begin{document}
\maketitle
\begin{abstract}

Recently, various intermediate layer distillation (ILD) objectives have been shown to improve compression of BERT models via Knowledge Distillation (KD).
However, a comprehensive evaluation of the objectives in both task-specific and task-agnostic settings is lacking.
To the best of our knowledge, this is the first work comprehensively evaluating distillation objectives in both settings.
We show that attention transfer gives the best performance overall.
We also study the impact of layer choice when initializing the student from the teacher layers, finding a significant impact on the performance in task-specific distillation. 
For vanilla KD and hidden states transfer, initialisation with lower layers of the teacher gives a considerable improvement over higher layers, especially on the task of QNLI (up to an absolute percentage change of 17.8 in accuracy). 
Attention transfer behaves consistently under different initialisation settings.
We release our code as an efficient transformer-based model distillation framework for further studies.\footnote{\href{https://github.com/mainlp/How-to-distill-your-BERT}{https://github.com/mainlp/How-to-distill-your-BERT}}
\end{abstract}

\section{Introduction}

Large-scale pre-trained language models (PLMs) have brought revolutionary advancements to natural language processing, such as BERT \cite{devlin-etal-2019-bert}, XLNet \cite{yang2019xlnet}, ELECTRA \cite{clark2020electra} and GPT-3 \cite{brown2020language}. 
However, the enormous size of these models has led to difficulties in deploying them in resource-constrained environments. Therefore significant interest has emerged in developing methods for reducing their size. 

Knowledge Distillation (KD) \cite{hinton2015distilling} transfers the knowledge embedded in one model to another, which can be used for cross-lingual transfer, cross-modal transfer, and model compression.
KD heavily depends on the distillation objective, which determines how knowledge is transferred.
Many works have tried to design different distillation objectives for Transformer-based \cite{vaswani2017attention} model compression and successfully distilled PLMs into smaller models, either task-specifically (\citealp{sun-etal-2019-patient}; \citealp{jiao-etal-2020-tinybert}) or task-agnostically---which differ in whether KD is performed at the pre-training stage or during task finetuning (\citealp{sanh2019distilbert}; \citealp{sun-etal-2020-mobilebert}; \citealp{wang2020minilm}; \citealp{wang-etal-2021-minilmv2}). 

Despite their impressive results, determining the best distillation objective is difficult due to their diverse comparison setups, such as data preprocessing, student model initialization, layer mapping strategies, task-specific/agnostic settings, and others. This breadth of choices and lack of code has led to comparison on unequal grounds and contradictory findings.\footnote{For example, both \citet{jiao-etal-2020-tinybert} and \citet{wang2020minilm} claimed to be the better method in their setting. See section \ref{section:objectives} for detail. } This shows a substantial need to reproduce and evaluate distillation objectives within the same setting. 
Motivated by this gap, we conduct experiments on the most common distillation objectives and their combinations in task-specific and task-agnostic settings. 
From our empirical evaluation, we show: (1) attention transfer performs consistently well in various initialisation settings, (2)  initialisation with lower layers of the teacher gives a considerable improvement over higher layers in task-specific distillation.

In summary, our \textbf{contributions} are:
\begin{itemize}
    \item We perform an evaluation of the effectiveness of different distillation objectives and the layer choice for initializing the student from the teacher layer.
    \item We make our code available as an efficient distillation framework.
    \item We provide practical guidance in terms of teacher layer choice for initialisation, distillation objectives and training parameters.
    
\end{itemize}

\section{Related Work}

\paragraph{Task-specific Distillation}
\citet{sun2019patient} task-specifically compressed BERT by learning from the every $k$-th layer of the teacher. 
To avoid leaving out some of the teacher layers, many follow-up works (\citealp{wu-etal-2020-skip}, \citealp{passban2021alp}, \citealp{wu-etal-2021-universal}) designed new layer mapping strategies to fuse the teacher layers. 
\citet{jiao-etal-2020-tinybert} used data augmentation to further improve the performance.
Initialising the student model with pre-trained weights is crucial for performance since the student learns from the teacher only shortly in downstream tasks.
Common choices for initialization are: (1) task-agnostically distilling models first, (2) using publicly available distilled models, or (3) initializing with teacher layers. 
As part of this study, we examine how to maximize the benefits of initializing from teacher layers. 

\paragraph{Task-agnostic Distillation}
In the field of task-agnostic distillation, one line of work is to compress the teacher model into a student model with the same depth but narrower blocks (\citealp{sun-etal-2020-mobilebert}, \citealp{zhang2022autodistill}).
Another line of work is to distill the teacher into a student with fewer layers (\citealp{sanh2019distilbert}, \citealp{jiao-etal-2020-tinybert}, \citealp{wang2020minilm}, \citealp{wang-etal-2021-minilmv2}), which is our focus.

\paragraph{Comparative Studies} 
\citet{li-etal-2021-select-one} conducted out-of-domain and adversarial evaluation on three KD methods, which used hidden state transfer or data augmentation.
\citet{lu2022knowledge} is closely related to our work, where they also evaluated knowledge types and initialisation schemes.
However, they did not consider layer choice when initialising from the teacher, and the evaluation was only for task-specific settings.
Hence, our work complements theirs.

\section{Distillation Objectives}

\paragraph{Prediction Layer Transfer}
Prediction layer transfer minimizes the soft cross-entropy between the logits from the teacher and the student: 
    $ \mathcal{L}_{\text{pred}}=\operatorname{CE}\left(\bm{z}^{T} / t, \bm{z}^{S} / t\right)$,
with $\bm{z}^T$ and $\bm{z}^S$ the logits from the teacher/student and $t$ is the temperature value.

Following the vanilla KD approach \cite{hinton2015distilling}, the final training loss is a combination of $\mathcal{L}_{\text{pred}}$ and supervision loss $\mathcal{L}_{\text{ce}}$ (masked language modelling loss $\mathcal{L}_{\text{mlm}}$ in the pertaining stage). We denote this objective as \textbf{vanilla KD}.

\paragraph{Hidden States Transfer}
Hidden states transfer penalizes the distance between the hidden states of specific layers from the teacher and the student.
Common choices for the representation are the embedding of the $\texttt{[CLS]}$ token \cite{sun2019patient} and the whole sequence embedding \cite{jiao-etal-2020-tinybert}.
We use Mean-Squared-Error (MSE) to measure the distance between the student and teacher embedding, which can be formulated as
$ \mathcal{L}_{\text {hid}}=\operatorname{MSE}\left(\boldsymbol{h}^S \boldsymbol{W}_h, \boldsymbol{h}^T\right)$, 
where $\pmb{h}^{S} \in \mathbb{R}^{d}$ and $\pmb{h}^{T} \in \mathbb{R}^{d^{\prime}}$ are the $\texttt{[CLS]}$ token embedding of specific student and teacher layer,
$d$ and $d^{\prime}$ are the hidden dimensions.
The matrix $\pmb{W}_{h} \in \mathbb{R}^{ d \times d^{\prime}}$ is a learnable transformation.
We denote this objective as \textbf{Hid-CLS}.
In the case of transferring the sequence embedding, one can replace the token embeddings with sequence embeddings $\pmb{H}^{S} \in \mathbb{R}^{ l \times d}$ and $\pmb{H}^{T} \in \mathbb{R}^{ l \times d^{\prime}}$, where $l$ is the sequence length. 
The objective that transfers the sequence embedding with MSE loss is denoted as \textbf{Hid-Seq}. 

We also evaluated a contrastive representation learning method which transfers the hidden state representation from the teacher to the student with a contrastive objective \cite{sun-etal-2020-contrastive}.
We inherited their code for implementation and refer our readers to the original paper for details. 
We denote this objective as \textbf{Hid-CLS-Contrast}.

 \paragraph{Attention and Value Transfer}
The attention mechanism has been found to capture rich linguistic knowledge \cite{clark-etal-2019-bert}, and attention map transfer is widely used in transformer model distillation.
To measure the similarity between the multi-head attention block of the teacher and the student, MSE and Kullback-Leibler divergence are the two standard loss functions.
The objective using MSE is formulated as
$\mathcal{L}_{\mathrm{att}}=\frac{1}{h} \sum_{i=1}^h \operatorname{MSE}(\boldsymbol{A}_i^S, \boldsymbol{A}_i^T)$, 
where $h$ is the number of attention heads, matrices $\boldsymbol{A}_{i} \in \mathbb{R}^{l \times l}$ refers to the $i$-th attention head (before the softmax operation) in the multi-head attention block.  
We denote this objective as \textbf{Att-MSE}.

\begin{table*}[ht!]
\small
    \centering
    \adjustbox{max width=\textwidth}{
    \begin{tabular}{l c c c c c c c c} 

    \toprule
    \textbf{Objectives} & \textbf{QNLI}  & \textbf{SST-2} & \textbf{MNLI} & \textbf{MRPC} & \textbf{QQP} & \textbf{RTE} & \textbf{CoLA} & \textbf {Avg}\\ 
                    &  Acc            &  Acc           & Acc           & F1            & Acc          & Acc          & Mcc                         \\
    \midrule
    Vanilla KD  & $\textrm{66.5}_{\pm \textrm{1.49}}$ & $\textrm{84.7}_{\pm \textrm{0.16}}$ & $\textrm{75.1}_{\pm \textrm{0.05}}$ & $\textrm{71.2}_{\pm \textrm{0.80}}$ & $\textrm{81.9}_{\pm \textrm{0.10}}$ & $\textrm{54.0}_{\pm \textrm{1.24}}$ & $\textrm{69.1}_{\pm \textrm{0.00}}$ & 71.8 \\
    \midrule
    Hid-CLS-Contrast  & $\textrm{69.3}_{\pm \textrm{0.60}}$ & $\textrm{85.3}_{\pm \textrm{0.56}}$ & $\textrm{76.2}_{\pm \textrm{0.45}}$ & $\textrm{71.1}_{\pm\textrm{0.85}}$ & $\textrm{83.1}_{\pm \textrm{0.69}}$ & $\textrm{53.6}_{\pm \textrm{0.23}}$ & $\textrm{69.0}_{\pm \textrm{0.12}}$  & 72.5\\
    Hid-CLS                      & $\textrm{75.7}_{\pm \textrm{0.57}}$ &$\textrm{85.8}_{\pm \textrm{0.34}}$ & $\textrm{77.0}_{\pm \textrm{0.10}}$ & $\textrm{71.3}_{\pm \textrm{0.41}}$   & $\textrm{83.8}_{\pm \textrm{1.63}}$ & $\textrm{54.0}_{\pm \textrm{2.17}}$ & $\textrm{68.4}_{\pm \textrm{0.35}}$  & 73.2 \\
    Hid-Seq                             & $\textrm{83.3}_{\pm \textrm{0.13}}$   & $\textrm{87.4}_{\pm \textrm{0.13}}$  & $\textrm{78.3}_{\pm \textrm{0.13}}$ & $\textbf{\textrm{72.9}}_{\pm \textrm{0.50}}$ & $\textrm{87.6}_{\pm \textrm{0.00}}$ & $\textrm{51.8}_{\pm \textrm{1.10}}$ & $\textrm{69.2}_{\pm \textrm{0.55}}$ & 75.8 \\
    \midrule
    Att-MSE                          & $\textrm{84.3}_{\pm \textrm{0.18}}$ & $\textrm{89.2}_{\pm \textrm{0.40}}$ & $\textrm{78.6}_{\pm \textrm{0.25}}$ & $\textrm{71.1}_{\pm \textrm{0.41}}$ & $\textrm{88.7}_{\pm \textrm{0.05}}$ & $\textrm{54.4}_{\pm \textrm{1.03}}$ & $\textrm{69.3}_{\pm \textrm{0.17}}$ & 76.5 \\
   \multicolumn{1}{l}{\:\:  +Hid-Seq}              & $\textrm{84.6}_{\pm \textrm{0.29}}$ & $\textrm{89.2}_{\pm \textrm{0.21}}$ & $\textrm{78.9}_{\pm \textrm{0.10}}$ & $\textrm{71.8}_{\pm \textrm{0.51}}$ & $\textrm{88.8}_{\pm \textrm{0.00}}$ & $\textrm{54.0}_{\pm \textrm{0.93}}$ & $\textbf{\textrm{69.5}}_{\pm \textrm{0.48}}$ & 77.0 \\
    \midrule  
    Att-KL                           & $\textrm{85.3}_{\pm \textrm{0.14}}$ & $\textrm{89.0}_{\pm \textrm{0.26}}$ & $\textrm{79.4}_{\pm \textrm{0.08}}$ & $\textrm{71.4}_{\pm \textrm{0.29}}$ & $\textrm{89.0}_{\pm \textrm{0.05}}$ & $\textrm{55.5}_{\pm \textrm{2.05}}$ & $\textrm{69.3}_{\pm \textrm{0.13}}$ & 77.0 \\
    
    \multicolumn{1}{l}{\:\:  +Hid-Seq}               & $\textrm{84.6}_{\pm \textrm{0.21}}$ & $\textrm{89.1}_{\pm \textrm{0.46}}$& $\textrm{79.5}_{\pm \textrm{0.17}}$& $\textrm{72.4}_{\pm \textrm{0.39}}$ & $\textrm{89.0}_{\pm \textrm{0.06}}$ & $\textrm{57.2}_{\pm \textrm{0.86}}$ & $\textrm{69.3}_{\pm \textrm{0.21}}$ & 77.3\\
         
    \multicolumn{1}{l}{\:\:  +Val-KL}               & $\textbf{\textrm{85.5}}_{\pm \textrm{0.24}}$ & $\textbf{\textrm{89.6}}_{\pm \textrm{0.31}}$ & $\textbf{\textrm{79.6}}_{\pm \textrm{0.10}}$ & $\textrm{72.2}_{\pm \textrm{0.39}}$ & $\textbf{\textrm{89.1}}_{\pm \textrm{0.05}}$ & $\textbf{\textrm{57.5}}_{\pm \textrm{0.70}}$ & $\textrm{69.2}_{\pm \textrm{0.15}}$ & $\textbf{77.5}$\\
    \bottomrule
    \end{tabular}}
\caption{Task-specific distillation results on GLUE dev sets. Student models are initialised with every 4th layer of the teacher model. We report the average and standard deviation over 4 runs. Attention based objectives consistently outperform hidden states transfer and vanilla KD. }
\label{table:idl-specific}
\end{table*}
Since the attention after the softmax function is a distribution over the sequence, we can also use the KL-divergence to measure the distance:
$\mathcal{L}_{\mathrm{att}}=\frac{1}{T H} \sum_{t=1}^{T} \sum_{h=1}^{H} D_{K L}(a_{t, h}^{T} \| a_{t, h}^{S})$,
where $T$ is the sequence length and $H$ is the number of attention heads. 
We will denote this objective as \textbf{Att-KL}. 
In addition to attention transfer, value-relation transfer was proposed by \citet{wang2020minilm}, to which we refer our readers for details. 
Value-relation transfer objective will be denoted as \textbf{Val-KL}.

\begin{table*}[ht]

\centering
 \adjustbox{max width=0.7\textwidth}{
\begin{tabular}{l c c c c c c c c}
\toprule 
\textbf{Objectives} & \textbf{QNLI}  & \textbf{SST-2} & \textbf{MNLI} & \textbf{MRPC} & \textbf{QQP} & \textbf{RTE} & \textbf{CoLA} & \textbf {Avg}\\
                &  Acc            &  Acc           & Acc           & F1            & Acc          & Acc          & Mcc                         \\
\midrule
$\text{DistilBERT}^{\star}$ & 89.2 & 91.3 & 82.2 & 87.5 & 88.5 & 59.9 & 51.3 &  78.5\\ 
$\text{TinyBERT}^{\dagger}$ &  90.5 & 91.6 & 83.5 & 88.4 & 90.6 & 72.2 & 42.8 & 79.9 \\
$\text{MiniLM}^{\mathsection}$ & \textbf{91.0} & 92.0 & \textbf{84.0} & 88.4 & \textbf{91.0} & \textbf{71.5} & 49.2 & 81.0  \\
\midrule
$\text{Vanilla KD}^{\star}$ & 88.6 & 91.4 & 82.4  &  86.5 & 90.6 & 61.0 & \textbf{54.4} &  79.3  \\
\midrule
Hid-CLS & 86.5 & 90.6 & 79.3 & 73.0 & 89.7 & 61.0 & 33.9 & 73.4 \\
Hid-Seq     & 89.2 &  91.5 & 82.3 & 89.2 & 90.3 &  67.2 & 48.2 & 79.7\\ 
\midrule
Att-MSE  & 89.8 &  91.6 & 83.2 & 90.6 & 90.7 & 69.7 & 53.5 & \textbf{81.3}\\ 

\multicolumn{1}{r}{+$\text{Hid-Seq}^{\dagger}$}   &  89.7 & \textbf{92.4} & 82.8 & 90.4 & 90.8 & 68.6 & 52.8 & 81.1\\ 
\midrule
Att-KL  & 88.0 & 89.7 & 81.1 & 90.1 & 90.3 & 66.1 & 43.6 & 78.4 \\
\multicolumn{1}{l}{\:  +Hid-Seq}  & 88.9 &91.6 & 82.4 &90.0 & 90.5& 66.8 & 47.9 & 79.7\\
\multicolumn{1}{l}{\:  +$\text{Val-KL}^{\mathsection}$}  & 89.8 & 91.6 & 82.4 & \textbf{91.0} & 90.6 & 66.7 & 47.7 & 80.0\\

\bottomrule

\end{tabular}}
\caption{Task-agnostic distillation: Performance on GLUE dev sets of three existing distilled 6-layer Transformer models and our 6-layer students distilled. All the students are randomly initialised and distilled from $\text{BERT}_{\text{BASE}}$. We report the best fine-tuning result with grid search over learning rate and batch size. Att-MSE performs the best among all the objectives.}
\label{table:idl-agnostic}
\end{table*}

\section{Experimental Setup}
We evaluate our model on the General Language Understanding Evaluation (GLUE) benchmark \cite{wang-etal-2018-glue} tasks, including linguistic acceptability (CoLA), sentiment analysis (SST-2), semantic equivalence (MRPC, QQP), and natural language inference (MNLI, QNLI, RTE).

For task-specific distillation, we distill a fine-tuned $\text{RoBERTa}_\text{BASE}$ \cite{liu2019roberta} into a 3-layer transformer model on each GLUE task, using the Fairseq \cite{ott2019fairseq} implementation and the recommended hyperparameters presented in \citet{liu2019roberta}.
We follow the training procedure from TinyBERT to  perform \textit{intermediate layer} and \textit{prediction layer} distillation sequentially for 10 epochs each, freeing us from tuning the loss weights.
For intermediate layer distillation, the student learns from the same teacher's layers that were used for initialising the student.
In addition, we always initialise the embedding layer with the teacher's embedding layer.

For task-agnostic distillation, we distill the uncased version of $\text{BERT}_{\text{base}}$ into a 6-layer student model, based on the implementation by \citet{izsak-etal-2021-train}.
Here we perform last-layer knowledge transfer since we see no improvement when transferring multiple layers in our experiments.
We train the student model for 100k steps with batch size 1024, a peaking learning rate of 5e-4 and a maximum sequence length of 128.
The distilled student model is then fine-tuned on the GLUE datasets with grid search over batch size \{16, 32\} and learning rate \{1e-5, 3e-5, 5e-5, 8e-5\}.
We follow the original training corpus of BERT: English Wikipedia and BookCorpus \cite{zhu2015aligning}.

\begin{table*}[t]
\centering
\small
\begin{tabular}{l c c c c c c c c c}
\toprule 
\textbf{Objectives} & \textbf{Init.} & \textbf{QNLI}  & \textbf{SST-2} & \textbf{MNLI} & \textbf{MRPC} & \textbf{QQP} & \textbf{RTE} & \textbf{CoLA} & \textbf {Avg}\\
                &&  Acc            &  Acc           & Acc           & F1            & Acc          & Acc          & Mcc                         \\
\midrule
\multirow{3}{*}{Vanilla KD} & 
\multicolumn{1}{l}{4,8,12}  & $\textrm{66.5}_{\pm \textrm{1.49}}$ & $\textrm{84.7}_{\pm \textrm{0.16}}$ & $\textrm{75.1}_{\pm \textrm{0.05}}$ & $\textrm{71.2}_{\pm \textrm{0.80}}$ & $\textrm{81.9}_{\pm \textrm{0.10}}$ & $\textrm{54.0}_{\pm \textrm{1.24}}$ & $\textrm{69.1}_{\pm \textrm{0.00}}$ & 71.8\\
&\multicolumn{1}{l}{1,8,12}  & $\textrm{82.9}_{\pm \textrm{0.31}}$ & $\textrm{88.5}_{\pm \textrm{0.51}}$ & $\textrm{76.6}_{\pm \textrm{0.08}}$ & $\textrm{71.2}_{\pm \textrm{0.88}}$ & $\textrm{87.8}_{\pm \textrm{0.06}}$ & $\textrm{55.5}_{\pm \textrm{1.07}}$ & $\textrm{70.8}_{\pm \textrm{0.29}}$ & 76.2\\
&\multicolumn{1}{l}{1,2,3}   & $\textbf{\textrm{86.2}}_{\pm \textrm{0.35}}$ & $\textbf{\textrm{90.4}}_{\pm \textrm{0.28}}$ & $\textbf{\textrm{78.7}}_{\pm \textrm{0.18}}$ & $\textbf{\textrm{78.6}}_{\pm \textrm{0.18}}$ & $\textbf{\textrm{89.8}}_{\pm \textrm{0.05}}$ & $\textbf{\textrm{57.1}}_{\pm \textrm{1.46}}$ & $\textbf{\textrm{74.9}}_{\pm \textrm{0.54}}$ & \textbf{79.4} \\
\midrule
\multirow{3}{*}{Hid-CLS-Contrast} &
\multicolumn{1}{l}{4,8,12}  & $\textrm{69.3}_{\pm \textrm{0.60}}$ & $\textrm{85.3}_{\pm \textrm{0.56}}$ & $\textrm{76.2}_{\pm \textrm{0.45}}$ & $\textrm{71.1}_{\pm\textrm{0.85}}$ & $\textrm{83.1}_{\pm \textrm{0.69}}$ & $\textrm{53.6}_{\pm \textrm{0.23}}$ & $\textrm{69.0}_{\pm \textrm{0.12}}$  & 72.5 \\
&\multicolumn{1}{l}{1,8,12}  & $\textrm{82.9}_{\pm \textrm{0.36}}$ & $\textrm{88.6}_{\pm \textrm{0.29}}$ & $\textrm{77.0}_{\pm \textrm{0.58}}$ & $\textrm{72.8}_{\pm\textrm{0.61}}$ & $\textrm{88.0}_{\pm \textrm{0.13}}$ & $\textrm{55.4}_{\pm \textrm{0.75}}$ & $\textrm{70.4}_{\pm \textrm{0.30}}$  & 76.4\\
&\multicolumn{1}{l}{1,2,3}   & $\textbf{\textrm{86.1}}_{\pm \textrm{0.22}}$ & $\textbf{\textrm{89.6}}_{\pm \textrm{0.38}}$ & $\textbf{\textrm{79.0}}_{\pm \textrm{0.12}}$ & $\textbf{\textrm{73.9}}_{\pm\textrm{1.43}}$ & $\textbf{\textrm{90.1}}_{\pm \textrm{0.10}}$ & $\textbf{\textrm{55.1}}_{\pm \textrm{0.67}}$ & $\textbf{\textrm{71.1}}_{\pm \textrm{1.09}}$  & \textbf{77.8}\\
\midrule
\multirow{3}{*}{Hid-CLS}  & \multicolumn{1}{l}{4,8,12}  & $\textrm{75.7}_{\pm \textrm{0.57}}$ &$\textrm{85.8}_{\pm \textrm{0.34}}$ & $\textrm{77.0}_{\pm \textrm{0.10}}$ & $\textrm{71.3}_{\pm \textrm{0.41}}$   & $\textrm{83.8}_{\pm \textrm{1.63}}$ & $\textrm{54.0}_{\pm \textrm{2.17}}$ & $\textrm{68.4}_{\pm \textrm{0.35}}$  & 73.2 \\
&\multicolumn{1}{l}{1,8,12}  & $\textrm{83.4}_{\pm \textrm{0.15}}$ &$\textrm{88.1}_{\pm \textrm{0.38}}$ & $\textrm{77.7}_{\pm \textrm{0.10}}$ & $\textrm{71.9}_{\pm \textrm{0.10}}$   & $\textrm{88.6}_{\pm \textrm{0.06}}$ & $\textrm{56.1}_{\pm \textrm{0.88}}$ & $\textrm{71.5}_{\pm \textrm{0.40}}$  & 76.7 \\
&\multicolumn{1}{l}{1,2,3}   & $\textbf{\textrm{85.7}}_{\pm \textrm{0.05}}$ &$\textbf{\textrm{90.3}}_{\pm \textrm{0.29}}$ & $\textbf{\textrm{78.6}}_{\pm \textrm{0.14}}$ & $\textbf{\textrm{74.3}}_{\pm \textrm{1.00}}$   & $\textbf{\textrm{90.1}}_{\pm \textrm{0.00}}$ & $\textbf{\textrm{57.1}}_{\pm \textrm{1.37}}$ & $\textbf{\textrm{73.6}}_{\pm \textrm{0.24}}$  & \textbf{78.5} \\
\midrule
\multirow{3}{*}{Hid-Seq}    &
\multicolumn{1}{l}{4,8,12}  &$\textrm{83.3}_{\pm \textrm{0.13}}$   & $\textrm{87.4}_{\pm \textrm{0.13}}$  & $\textrm{78.3}_{\pm \textrm{0.13}}$ & $\textrm{72.9}_{\pm \textrm{0.50}}$ & $\textrm{87.6}_{\pm \textrm{0.00}}$ & $\textrm{51.8}_{\pm \textrm{1.10}}$ & $\textrm{69.2}_{\pm \textrm{0.55}}$ & 75.8 \\
&\multicolumn{1}{l}{1,8,12}  & $\textrm{84.3}_{\pm \textrm{0.10}}$   & $\textrm{88.6}_{\pm \textrm{0.28}}$  & $\textrm{78.2}_{\pm \textrm{0.08}}$ & $\textrm{72.0}_{\pm \textrm{0.70}}$ & $\textrm{88.6}_{\pm \textrm{0.10}}$ & $\textrm{55.2}_{\pm \textrm{1.40}}$ & $\textrm{71.6}_{\pm \textrm{0.37}}$ & 77.6\\
&\multicolumn{1}{l}{1,2,3}   &$\textbf{\textrm{85.9}}_{\pm \textrm{0.24}}$   & $\textbf{\textrm{90.7}}_{\pm \textrm{0.08}}$  & $\textbf{\textrm{78.9}}_{\pm \textrm{0.10}}$ & $\textbf{\textrm{75.5}}_{\pm \textrm{1.14}}$ & $\textbf{\textrm{90.0}}_{\pm \textrm{0.05}}$ & $\textbf{\textrm{56.6}}_{\pm \textrm{0.74}}$ & $\textbf{\textrm{74.2}}_{\pm \textrm{0.45}}$ & \textbf{78.8}\\
\midrule
\multirow{3}{*}{Att-KL}  &
\multicolumn{1}{l}{4,8,12} & $\textrm{85.3}_{\pm \textrm{0.14}}$ & $\textrm{89.0}_{\pm \textrm{0.26}}$ & $\textbf{\textrm{79.4}}_{\pm \textrm{0.08}}$ & $\textrm{71.4}_{\pm \textrm{0.29}}$ & $\textrm{89.0}_{\pm \textrm{0.05}}$ & $\textrm{55.5}_{\pm \textrm{2.05}}$ & $\textrm{69.3}_{\pm \textrm{0.13}}$ & 77.0\\
&\multicolumn{1}{l}{1,8,12} & $\textrm{84.7}_{\pm \textrm{0.26}}$ & $\textbf{\textrm{89.6}}_{\pm \textrm{0.13}}$ & $\textrm{78.2}_{\pm \textrm{0.10}}$ & $\textbf{\textrm{72.5}}_{\pm \textrm{0.24}}$ & $\textrm{88.6}_{\pm \textrm{0.08}}$ & $\textrm{56.5}_{\pm \textrm{0.44}}$ & $\textbf{\textrm{70.4}}_{\pm \textrm{0.26}}$ & 77.2 \\
&\multicolumn{1}{l}{1,2,3}  &$\textbf{\textrm{86.2}}_{\pm \textrm{0.06}}$ & $\textrm{88.6}_{\pm \textrm{0.19}}$ & $\textrm{77.9}_{\pm \textrm{0.17}}$ & $\textrm{71.3}_{\pm \textrm{0.24}}$ & $\textbf{\textrm{89.0}}_{\pm \textrm{0.05}}$ & $\textbf{\textrm{61.2}}_{\pm \textrm{0.72}}$ & $\textrm{69.5}_{\pm \textrm{0.80}}$ & \textbf{77.7} \\
\midrule
\multirow{3}{*}{Att-MSE} &  
\multicolumn{1}{l}{4,8,12} & $\textrm{84.3}_{\pm \textrm{0.18}}$ & $\textrm{89.2}_{\pm \textrm{0.40}}$ & $\textbf{\textrm{78.6}}_{\pm \textrm{0.25}}$ & $\textrm{71.1}_{\pm \textrm{0.41}}$ & $\textrm{88.7}_{\pm \textrm{0.05}}$ & $\textrm{54.4}_{\pm \textrm{1.03}}$ & $\textrm{69.3}_{\pm \textrm{0.17}}$ & 76.5 \\
&\multicolumn{1}{l}{1,8,12} & $\textrm{84.3}_{\pm \textrm{0.25}}$ & $\textbf{\textrm{89.8}}_{\pm \textrm{0.39}}$ & $\textrm{77.5}_{\pm \textrm{0.14}}$ & $\textbf{\textrm{72.5}}_{\pm \textrm{1.36}}$ & $\textrm{88.4}_{\pm \textrm{0.05}}$ & $\textrm{57.2}_{\pm \textrm{0.96}}$ & $\textbf{\textrm{70.6}}_{\pm \textrm{0.45}}$ & 77.2\\
&\multicolumn{1}{l}{1,2,3} &  $\textbf{\textrm{86.2}}_{\pm \textrm{0.13}}$ & $\textrm{88.2}_{\pm \textrm{0.43}}$ & $\textrm{77.8}_{\pm \textrm{0.13}}$ & $\textrm{72.4}_{\pm \textrm{0.49}}$ & $\textbf{\textrm{88.8}}_{\pm \textrm{0.00}}$ & $\textbf{\textrm{60.3}}_{\pm \textrm{1.49}}$ & $\textrm{69.6}_{\pm \textrm{0.90}}$ & \textbf{77.6}\\
\bottomrule
\end{tabular}
\caption{Task-specific distillation: Performance of the student initialised with different teacher layers over 4 runs. For vanilla KD and Hid-CLS transfer, the performance on QNLI is significantly improved when initialising with lower teacher layers. Attention transfer benefits less from initialising from lower teacher layers.}
\label{tab:specific-init}
\end{table*}

\section{Results}
\paragraph{Distillation Objectives}
\label{section:objectives}
Distillation objective performances are compared in Table \ref{table:idl-specific} and Table \ref{table:idl-agnostic} for task-specific and task-agnostic settings, respectively.
In the task-specific setting, attention transfer is the best choice with initialisation from every \textit{k}-th teacher layer.
However, the performance of hidden states transfer and \textit{vanilla KD} can be drastically improved under other initialisation settings, which we discuss in the next section.

In the task-agnostic setting, the \textit{Att-MSE} objective outperforms \textit{Att-KL}, which performs similarly to \textit{vanilla KD} and hidden states transfer.  
This contradicts the observation in MiniLM \cite{wang2020minilm}, where their \textit{Att-KL} based objective outperforms TinyBERT \cite{jiao-etal-2020-tinybert} with \textit{Att-MSE}. 
However, MiniLM has more training iterations and a larger batch size, which makes comparison difficult.
The performance drop of \textit{Att-KL} compared to \textit{Att-MSE} is mainly due to its poor performance on CoLA (linguistic acceptability of a sentence), on which MiniLM also performs poorly.
We hypothesise that MSE can transfer the linguistic knowledge embedded in the attention matrix more effectively because the MSE loss function gives more direct matching than KL-divergence, which was also concluded by \citet{kim2021comparing}. 

For reference, we report the result of 3 existing works that use the same objectives in our experiments.
The result of DistilBERT and MiniLM are taken from the respective papers.
The result of TinyBERT is taken from \citet{wang2020minilm} for fair comparison since TinyBERT only reported task-specific distillation result with data augmentation.
We denote the prior works and the corresponding objective we evaluate with the same superscript symbol.

\begin{table*}[t]
\small
\centering
\begin{tabular}{l l c c c c c c c c}
\toprule 
\textbf{Objectives} & \textbf{Init.} & \textbf{QNLI}  & \textbf{SST-2} & \textbf{MNLI} & \textbf{MRPC} & \textbf{QQP} & \textbf{RTE} & \textbf{CoLA} & \textbf {Avg}\\
&               &  Acc            &  Acc           & Acc           & F1            & Acc          & Acc          & Mcc                         \\
\midrule
\multirow{2}{*}{Vanilla KD}    
& random  & \textbf{88.6} & \textbf{91.4} & \textbf{82.4}  &  86.5 & 90.6 & 61.0 & 54.4  & 79.3 \\
& first 6   &  88.3    & 91.2 & 82.2    &  \textbf{87.0}   & 90.6    & \textbf{62.8} & \textbf{55.4} &  \textbf{79.6}\\
\midrule
\multirow{2}{*}{Hid-CLS}    
& random  & 86.5 & 90.6 & 79.3 & 73.0 & 89.7 & 61.0 & 33.9& 73.4  \\
& first 6   & \textbf{87.0}    & \textbf{91.2} & \textbf{80.7}    &  \textbf{88.0}   & \textbf{90.2}    & \textbf{66.0} & \textbf{42.5} &  \textbf{77.9}\\
\midrule
\multirow{2}{*}{Hid-Seq}     
& random  & \textbf{\text{89.2}} &  91.5 & 82.3 & 89.2 & 90.3 &  \textbf{67.2} & 48.2 & 79.7  \\
& first 6   &  87.5    & 91.5 & 82.3    &  \textbf{90.0}   &   \textbf{90.5}  & 66.4 & \textbf{50.6} & \textbf{79.9} \\
\midrule
\multirow{2}{*}{Att-MSE}  
& random & \textbf{89.8} & 91.6 & \textbf{83.2} & 90.6 & 90.7 &\textbf{69.7} & \textbf{53.5} & \textbf{81.3}\\
& first 6 & 89.5 & \textbf{91.7} &  82.8    & \textbf{91.0} & \textbf{90.8}  & 66.1 & 53.4 & 80.8\\
\bottomrule
\end{tabular}

\caption{Task-agnostic distillation:  Performance of the student initialised with random weights vs first 6 teacher layers. Attention transfer performs the best in both initialisation settings.}
\label{tab:agnostic-init}
\end{table*}
\paragraph{Initialisation}
We also studied the impact of the choice of  teacher layers for initialising the student. 
Evaluation score on GLUE task development sets under different teacher layer choices for initialisation are reported in Table \ref{tab:specific-init} and Table \ref{tab:agnostic-init} for task-specific and task-agnostic distillation, respectively.

We observe that initiatlization of layers has a huge impact in the task-specific setting.
The performance of \textit{vanilla KD} and Hidden states transfer was significantly improved when initialising from lower layers of the teacher (e.g.\ from 68.1\% to 85.9\% on QNLI for Vanilla KD).
This explains the impressive result of PKD \cite{sun2019patient}, which initialised the student with first k teacher layers.
We believe this is an important observation that will motivate further research into investigating the effectiveness of the different layers of the pre-trained transformer model. 

In the task-agnostic setting,  we only observe considerable improvement with the objective \textit{Hid-CLS}, which performs poorly when randomly initialized, compared to other objectives.
This contradicts \citet{sanh2019distilbert} with a \textit{vanilla KD} objective, where they instead showed improvement of 3 average score when initialising from the teacher over random initialisation.
However, our \textit{vanilla-KD} approach initialised with random weights outperforms their best result (79.3 vs 78.5).
Therefore, we hypothesise that the advantage of pre-loading teacher layers over random initialisation diminishes as the student is fully distilled during pre-training.

\paragraph{Significance Test} 
We conducted paired t-testing for all the distillation objectives in Table \ref{table:idl-specific} and the three initialisation choices within each objective in Table \ref{tab:specific-init}.
For Table \ref{table:idl-specific}, all the pairs of objectives are statistically significant (p < 0.05) except four: (Att-KL, Att-MSE), (Att-KL, Att-KL + Hid-Seq), (Att-KL, Att-MSE + Hid-Seq), (Att-MSE, Att-MSE + Hid-Seq).
This further supports our conclusion that when initialised from every K teacher layer, it is important to do attention transfer, and the specific objective matters less.
For Table \ref{tab:specific-init}, all three initialisation choices are statistically significantly different from each other for all the objectives, except the pair (1,8,12, 1,2,3) for Att-KL and Att-MSE, which indicates the robustness of attention transfer under different initialisation choices.

\paragraph{Training Time}
Since task-agnostic distillation is computationally expensive, we also focus on optimizing our distillation framework for faster training.
Our training time is about 58 GPU hours on 40GB A100, compared to TinyBERT (576 GPU hours on 16GB V100) and DistilBERT (720 GPU hours on 16GB V100). 
This is achieved by using a shorter sequence length and an optimized transformer pre-training framework by \citet{izsak-etal-2021-train}.
We see no improvement when using a longer sequence length of 512.

\paragraph{Guidance} To sum up, our observations, trade-offs and recommendations are: 
\begin{itemize}
\setlength\itemsep{0em}
\item For task-specific KD, we recommend attention transfer in general, due to its consistently high performance in various initialisation settings (Table \ref{tab:specific-init}). 
The exact attention distillation objective matter less (Table \ref{table:idl-specific}).
Considering the excellent performance of the vanilla KD approach (Table \ref{tab:specific-init}) when initialising with lower teacher layers, we also recommend lower teacher layer initialisation with the vanilla KD approach for its shorter training time and simple implementation.
\item For task-agnostic KD, attention transfer with Mean-Squared-Error is the best choice based on our result (Table \ref{table:idl-agnostic}, \ref{tab:agnostic-init}). 
\item  We recommend readers to use our task-agnostic distillation framework and short sequence length  for fast training.
\end{itemize}

\section{Conclusion}
We extensively evaluated distillation objectives for the transformer model and studied the impact of weight initialisation. 
We found that attention transfer performs consistently well in both task-specific and task-agnostic settings, regardless of the teacher layers chosen for student initialization.
We also observed that initialising with lower teacher layers significantly improved task-specific distillation performance compared to higher layers.
We release our code and hope this work motivates further research into developing better distillation objectives and compressing in-house models.

\section{Limitations}
We evaluated the most widely used distillation objectives including prediction layer transfer, hidden states transfer and attention transfer.
However, some objectives are not included in our evaluation due to missing implementation details in their paper.
For example, we only implemented the contrastive intermediate layer distillation objective proposed by \citet{sun-etal-2020-contrastive} in task-specific setting, since code and implementation details are missing for task-agnostic setting.
New objectives are increasingly appearing for model compression in the field of computer vision, such as Wasserstein contrastive representation distillation \cite{chen2021wasserstein} and distillation with Pearson correlation \cite{huang2022knowledge}, which can be included to have a broader scope of distillation objectives evaluation. 

This work empirically studied the impact of the teacher layer choice for initialization and training objectives, however, further analysis is needed to understand why lower teacher layers are essential for initialisation, and why attention transfer behaves consistently well under various teacher layer choices in the task-specific setting, while hidden state transfer does not.
\section*{Acknowledgements}
We thank the anonymous reviewers as well as the
members of the MaiNLP research lab for their
constructive feedback. 
This research is supported by ERC Consolidator
Grant DIALECT 101043235.

\bibliography{anthology,custom}
\bibliographystyle{acl_natbib}

\newpage
\appendix

\section{Hyperparameters}
\label{sec:hyper}

Table \ref{tab:config} shows the hyperparameters we use for task-agnostic distillation.

\begin{table}[ht]
    \centering
    \begin{adjustbox}{width={0.35\textwidth},keepaspectratio}%

    \begin{tabular}{l r}
        \toprule 
        \textbf{Hyperparameter} & \textbf{Our Model} \\
        \midrule
        Number of Layers & 6 \\
        Hidden Size & 768  \\
        FFN inner hidden size & 3072 \\
        Attention heads & 12 \\
        Attention head size & 64 \\
        Learning Rate Decay & Linear \\
        Weight Decay & 0.01 \\
        Optimizer & AdamW \\
        Adam $\epsilon$ & 1e-6 \\
        Adam $\beta_{1}$ & 0.9 \\
        Adam $\beta_{2}$ & 0.99 \\
        Gradient Clipping & 0.0 \\
        Warmup Proportion & 6\% \\
        Peak Learning Rate & 5e-4 \\
        Batch size & 1024 \\
        Max Steps & 100k\\
        \bottomrule
    \end{tabular}
    \end{adjustbox}

    \caption{Hyperparameter used for distilling our student model in the pre-training stage. }
    \label{tab:config}
\end{table}
\begin{table}[ht]
    \centering
    \begin{adjustbox}{width={0.4\textwidth},totalheight={\textheight},keepaspectratio}%

    \begin{tabular}{c c}
        \toprule 
        \textbf{Hyperparameter} & \textbf{Search Space} \\
        \midrule
        Learning Rate & \{1e-5, 3e-5, 5e-5, 8e-5\} \\
        Batch Size & \{16, 32\}  \\
        \bottomrule
    \end{tabular}
    \end{adjustbox}

    \caption{The hyperparameter space used for fine-tuning our distilled student model on GLUE benchmark tasks. }
    \label{tab:search}
\end{table}

As the distillation in the pre-training stage is computationally expensive and unstable, we suggest readers to follow our settings to avoid additional costs.
For example, we observed training loss divergence when using a higher learning rate (1e-3).

Table \ref{tab:search} shows the search space of learning rate and batch size for fine-tuning the general-distilled student.
We finetune for 10 epochs on each GLUE task.

For task-specific distillation, we follow the suggested hyperparameters shown in  the \href{https://github.com/facebookresearch/fairseq/tree/main/examples/roberta/config/finetuning}{repository} of RoBERTa \cite{liu2019roberta}. 

\begin{table*}[!b]
\centering
 \adjustbox{max width=1\textwidth}{
\begin{tabular}{l c c c c c c c c}
\toprule 
 & \textbf{Iteration Steps}  & \textbf{Batch Size} & \textbf{Layer Matching} & \textbf{Initialisation} & \textbf{Max Sequence Length} & \textbf{GPU hours} & \textbf{Avg-score}\\
\midrule
DistilBERT &-& 4k & prediction layer & every second teacher layer & 512 & 720h on 16GB V100 & 78.5\\ 
TinyBERT &   -  & 256& every second hidden layer & random & 128 & 576h on 16GB $\textrm{V100}^{\star}$ & 79.9\\
MiniLM & 400k  & 1024 & last hidden layer & random & 512 & - &  81.0\\
Ours & 100k & 1024 & last hidden layer & random & 128&58h on 40GB A100 & \textbf{81.3}\\
\bottomrule
\end{tabular}}
\caption{ Comparison of hyperparameter choices and training time between ours and prior works. Empty entries indicate that the papers do not report those numbers. $\star$: Number according to their GitHub issue answer.}
\label{table:setting-compare}
\end{table*}

\section{Comparison to prior works}
\label{sec:compare_exist}
Table \ref{table:setting-compare} compares the settings and computational costs of three prior works: DistilBERT \cite{sanh2019distilbert}, TinyBERT \cite{jiao-etal-2020-tinybert} and MiniLM \cite{wang2020minilm}, with our best-performing objective.
There are some differences between our settings and theirs, such as layer matching strategies (which teacher layers to transfer), initialisation choices, training steps and batch size. 
Comparatively, our framework requires less training time and can achieve comparable or better results.
Our training takes 58 GPU hours on A100 compared to 720 GPU hours on V100 for training DistilBERT (taking into consideration that an A100 GPU is about twice as fast as a V100).

\end{document}